\documentclass[sigconf]{acmart}
\usepackage{booktabs, tabularx, amsmath, mathtools, fancyhdr, balance} 
\usepackage[skip=4pt,font=footnotesize]{caption}
\usepackage{xargs}
\usepackage{xcolor}
\usepackage{color}
\usepackage[colorinlistoftodos,prependcaption,textsize=small]{todonotes}
\newcommandx{\unsure}[2][1=]{\todo[linecolor=red,backgroundcolor=red!25,bordercolor=red,#1]{#2}}
\newcommandx{\change}[2][1=]{\todo[linecolor=blue,backgroundcolor=blue!25,bordercolor=blue,#1]{#2}}
\newcommandx{\improve}[2][1=]{\todo[linecolor=green,backgroundcolor=green!25,bordercolor=green,#1]{#2}}
\newcommandx{\info}[2][1=]{\todo[linecolor=blue,backgroundcolor=blue!15,bordercolor=blue,#1]{#2}}

\setcopyright{rightsretained}

\acmDOI{}

\acmISBN{}

\acmConference[KDD '18]{ACM Knowledge Discovery and Data Mining}{August 2018}{London, United Kingdom} 
\acmYear{2018}
\copyrightyear{2018}

\begin{document}
\title[Detecting Spacecraft Anomalies Using LSTMs and Nonparametric Thresholding]{Detecting Spacecraft Anomalies Using LSTMs and Nonparametric Dynamic Thresholding}

\copyrightyear{2018} 
\acmYear{2018} 
\setcopyright{usgovmixed}
\acmConference[KDD '18]{The 24th ACM SIGKDD International Conference on Knowledge Discovery \& Data Mining}{August 19--23, 2018}{London, United Kingdom}
\acmBooktitle{KDD '18: The 24th ACM SIGKDD International Conference on Knowledge Discovery \& Data Mining, August 19--23, 2018, London, United Kingdom}
\acmPrice{15.00}
\acmDOI{10.1145/3219819.3219845}
\acmISBN{978-1-4503-5552-0/18/08}


\author{Kyle Hundman}
\affiliation{%
  \institution{NASA Jet Propulsion Laboratory}
  \institution{California Institute of Technology}
}
\email{kyle.a.hundman@jpl.nasa.gov}

\author{Valentino Constantinou}
\affiliation{%
  \institution{NASA Jet Propulsion Laboratory}
  \institution{California Institute of Technology}
}
\email{vconstan@jpl.nasa.gov}

\author{Christopher Laporte}
\affiliation{%
  \institution{NASA Jet Propulsion Laboratory}
  \institution{California Institute of Technology}
}
\email{christopher.d.laporte@jpl.nasa.gov}

\author{Ian Colwell}
\affiliation{%
  \institution{NASA Jet Propulsion Laboratory}
  \institution{California Institute of Technology}
}
\email{ian.colwell@jpl.nasa.gov}

\author{Tom Soderstrom}
\affiliation{%
  \institution{NASA Jet Propulsion Laboratory}
  \institution{California Institute of Technology}
}
\email{tom.soderstrom@jpl.nasa.gov}

\renewcommand{\shortauthors}{Hundman, Constantinou, Laporte, Colwell, Soderstrom}

\begin{abstract}

As spacecraft send back increasing amounts of telemetry data, improved anomaly detection systems are needed to lessen the monitoring burden placed on operations engineers and reduce operational risk. Current spacecraft monitoring systems only target a subset of anomaly types and often require costly expert knowledge to develop and maintain due to challenges involving scale and complexity. We demonstrate the effectiveness of Long Short-Term Memory (LSTMs) networks, a type of Recurrent Neural Network (RNN), in overcoming these issues using expert-labeled telemetry anomaly data from the Soil Moisture Active Passive (SMAP) satellite and the Mars Science Laboratory (MSL) rover, Curiosity. We also propose a complementary unsupervised and nonparametric anomaly thresholding approach developed during a pilot implementation of an anomaly detection system for SMAP, and offer false positive mitigation strategies along with other key improvements and lessons learned during development. 
\end{abstract}

%
%



\maketitle

\section{Introduction}
\label{Introduction}

Spacecraft are exceptionally complex and expensive machines with thousands of telemetry channels detailing aspects such as temperature, radiation, power, instrumentation, and computational activities.  Monitoring these channels is an important and necessary component of spacecraft operations given their complexity and cost. In an environment where a failure to detect and respond to potential hazards could result in the full or partial loss of spacecraft, anomaly detection is a critical tool to alert operations engineers of unexpected behavior. 


Current anomaly detection methods for spacecraft telemetry primarily consist of tiered alarms indicating when values stray outside of pre-defined limits and manual analysis of visualizations and aggregate channel statistics. Expert systems and nearest neighbor-based approaches have also been implemented for a small number of spacecraft \cite{fuertes2016improving}. These approaches have well-documented limitations -- extensive expert knowledge and human capital are needed to define and update nominal ranges and perform ongoing analysis of telemetry. Statistical and limit-based or density-based approaches are also prone to missing anomalies that occur within defined limits or those characterized by a temporal element \cite{Chandola2009}.  




These issues will be exacerbated as improved computing and storage capabilities lead to increasing volumes of telemetry data. NISAR, an upcoming Synthetic Aperture Radar (SAR) satellite, will generate around 85 terabytes of data per day and represents exponentially increasing data rates for Earth Science satellites \cite{nasa_2018}. Mission complexity and condensed mission time frames also call for improved anomaly detection solutions. For instance, the Europa Lander concept would have an estimated 20-40 days on Europa's surface due to high radiation and would require intensive monitoring during surface operations \cite{hand2017report}. Anomaly detection methods that are more accurate and scalable will help allocate limited engineering resources associated with such missions.

Challenges central to anomaly detection in multivariate time series data also hold for spacecraft telemetry. A lack of labeled anomalies necessitates the use of unsupervised or semi-supervised approaches. Real-world systems are usually highly non-stationary and dependent on current context. Data being monitored are often heterogeneous, noisy, and high-dimensional. In scenarios where anomaly detection is being used as a diagnostic tool, a degree of interpretability is required. Identifying the existence of a potential issue on board a spacecraft without providing any insight into its nature is of limited value to engineers. Lastly, a suitable balance must be found between the minimization of false positives and false negatives according to a given scenario.


\textbf{Contributions.} In this paper, we adapt and extend methods from various domains to mitigate and balance the issues mentioned above. This work is presented through the lens of spacecraft anomaly detection, but applies generally to many other applications involving anomaly detection for multivariate time series data. Specifically, we describe our use of Long Short-Term Memory (LSTM) recurrent neural networks (RNNs) to achieve high prediction performance while maintaining interpretability throughout the system. Once model predictions are generated, we offer a nonparametric, dynamic, and unsupervised thresholding approach for evaluating residuals. This approach addresses diversity, non-stationarity, and noise issues associated with automatically setting thresholds for data streams characterized by varying behaviors and value ranges. Methods for utilizing user-feedback and historical anomaly data to improve system performance are also detailed.

We then present experimental results using real-world, expert-labeled data derived from Incident Surprise, Anomaly (ISA) reports for the Mars Science Laboratory (MSL) rover, Curiosity, and the Soil Moisture Active Passive (SMAP) satellite. These reports are used by mission personnel to process unexpected events that impact a spacecraft and place it in potential risk during post-launch operations. Lastly, we highlight key milestones, improvements, and observations identified through an early implementation of the system for the SMAP mission and offer open source versions of methodologies and data for use by the broader research community\footnote{https://github.com/khundman/telemanom}. 

\section{Background and Related Work}
\label{Background and Related Work}

The breadth and depth of research in anomaly detection offers numerous definitions of anomaly types, but with regard to time-series data it is useful to consider three categories of anomalies -- \textit{point}, \textit{contextual}, and \textit{collective} \cite{Chandola2009}. \textit{Point} anomalies are single values that fall within low-density regions of values, \textit{collective} anomalies indicate that a sequence of values is anomalous rather than any single value by itself, and \textit{contextual} anomalies are single values that do not fall within low-density regions yet are anomalous with regard to local values. We use these characterizations to aid in comparisons of anomaly detection approaches and further profile spacecraft anomalies from SMAP and MSL.

Utility across application domains, data types, and anomaly types has ensured that a wide variety of anomaly detection approaches have been studied \cite{Chandola2009,Goldstein2016}. Simple forms of anomaly detection consist of out-of-limits (OOL) approaches which use predefined thresholds and raw data values to detect anomalies. A myriad of other anomaly detection techniques have been introduced and explored as potential improvements over OOL approaches, such as clustering-based approaches \cite{Iverson04inductivesystem,Gao2012,Li2016}, nearest neighbors approaches \cite{Bay2003,Iverson2008,LocalOutlierFactor,LocalOutlierProbabilities}, expert systems \cite{Tallo1992, ExpertSystemEnvironmentallyInducedAnomalies,Intelsat,Nozomi}, and dimensionality reduction approaches \cite{Scholkopf1998, Fujimaki2005, TakehisaYairi}, among others. These approaches represent a general improvement over OOL approaches and have been shown to be effective in a variety of use cases, yet each has its own disadvantages related to parameter specification, interpretability, generalizability, or computational expense \cite{Chandola2009, Goldstein2016} (see \cite{Chandola2009} for a survey of anomaly detection approaches). Recently, RNNs have demonstrated state-of-the-art performance on a variety of sequence-to-sequence learning benchmarks and have shown effectiveness across a variety of domains \cite{Schmidhuber2015}. In the following sections, we discuss the shortcomings of prior approaches in aerospace applications and demonstrate RNN's capacity to help address these challenges.

\subsection{Anomaly Detection in Aerospace}


Numerous anomaly detection approaches mentioned in the previous section have been applied to spacecraft. Expert systems have been used with numerous spacecraft \cite{Tallo1992, Ciceri1994, ExpertSystemEnvironmentallyInducedAnomalies, Intelsat}, notably the ISACS-DOC (Intelligent Satellite Control Software DOCtor) with the Hayabusa, Nozomi, and Geotail missions \cite{Nozomi}. Nearest neighbor based approaches have been used repeatedly to detect anomalies on board the Space Shuttle and the International Space Station \cite{Bay2003,Iverson2008}, as well as the XMM-Newton satellite \cite{NoveltyDetection}. The Inductive Monitoring System (IMS), also used by NASA on board the Space Shuttle and International Space Station, employs the practitioner's choice of clustering technique in order to detect anomalies, with anomalous observations falling outside of well-defined clusters \cite{Iverson04inductivesystem,Iverson2008}. ELMER, or Envelope Learning and Monitoring using Error Relaxation, attempts to periodically set new OOL bounds estimated using a neural network, aiming to reduce false positives and improve the performance of OOL anomaly detection tasks aboard the Deep Space One spacecraft \cite{Bernard1999}.

The variety of prior anomaly detection approaches applied to spacecraft would suggest their wide-spread use, yet out-of-limits (OOL) approaches remain the most widely used forms of anomaly detection in the aerospace industry \cite{NoveltyDetection,QuanLi2010,TakehisaYairi}. Despite their limitations, OOL approaches remain popular due to numerous factors -- low computational expense, broad and straight-forward applicability, and ease of understanding -- factors which may not all be present in more complex anomaly detection approaches. NASA's Orca and IMS tools, which employ nearest neighbors and clustering approaches, successfully detected all anomalies identified by Mission Evaluation Room (MER) engineers aboard the STS-115 mission (high recall) but also identified many non-anomalous events as anomalies (low precision), requiring additional work to mitigate against excessive false positives \cite{Iverson2008}. The IMS, as a clustering-based approach, limits representation of prior data to four coarse statistical features: average, standard deviation, maximum, and
minimum, and requires careful parameterization of time windows \cite{NoveltyDetection}. As a neural network, ELMER was only used for 10 temperature sensors on Deep Space One due to limitations in on-board memory and computational resources \cite{DeepSpaceOne}. Notably, none of these approaches make use of data beyond prior telemetry values.

For other missions considering the previous approaches, the potential benefits are often not enough to outweigh their limitations and perceived risk. This is partially attributable to the high complexity of spacecraft and the conservative nature of their operations, but these approaches have not demonstrated results and generalizability compelling enough to justify widespread adoption.  OOL approaches remain widely utilized because of these factors, but this is poised to change as data volumes grow and as RNN approaches demonstrate profound improvements in similar domains and applications.

\subsection{Anomaly Detection using LSTMs} 


The recent advancement of deep learning, compute capacity, and neural network architectures have lead to performance breakthroughs for a variety of problems including sequence-to-sequence learning tasks \cite{Graves2012,Graves2013,Sutskever2014}. Until recently, previous applications in aerospace involving large sets of high-dimensional data were forced to use methods less capable of modeling temporal information. Specifically, LSTMs and related RNNs represent a significant leap forward in efficiently processing and prioritizing historical information valuable for future prediction. When compared to dense Deep Neural Networks (DNN) and early RNNs, LSTMs have been shown to improve the ability to maintain memory of long-term dependencies due to the introduction of a weighted self-loop conditioned on context that allows them to forget past information in addition to accumulating it \cite{goodfellow2016deep,Sak2014, Malhorta2015}. Their ability to handle high-complexity, temporal or sequential data has ensured their widespread application in domains including natural language processing (NLP), text classification, speech recognition, and time series forecasting, among others \cite{Young2017,Zhou2016,Sak2014,Malhorta2015}. 

The inherent properties of LSTMs makes them an ideal candidate for anomaly detection tasks involving time-series, non-linear numeric streams of data. LSTMs are capable of learning the relationship between past data values and current data values and representing that relationship in the form of learned weights \cite{HochreiterLSTM,Bontemps2017}. When trained on nominal data, LSTMs can capture and model normal behavior of a system \cite{Bontemps2017}, providing practitioners with a model of system behavior under normal conditions. They can also handle multivariate time-series data without the need for dimensionality reduction \cite{Nanduri2016AnomalyDI} or domain knowledge of the specific application \cite{Taylor2016}, allowing for generalizability across different types of spacecraft and application domains. In addition, LSTM approaches have been shown to model complex nonlinear feature interactions \cite{Ogunmolu2016} that 
are often present in multivariate time-series data streams, and obviate the need to specify a time-window in which to consider data values in an anomaly detection task due to the use of shared parameters across time \cite{Malhorta2015,goodfellow2016deep}.

These advantages have motivated the use of LSTM networks in several recent anomaly detection tasks \cite{Taylor2016,Chauhan2015,Nanduri2016AnomalyDI, Malhorta2015,Malhotra2016LSTMbasedEF,Bontemps2017}, where LSTM models are fit on nominal data and model predictions are compared to actual data stream values using a set of detection rules in order to detect anomalies \cite{Malhorta2015,Malhotra2016LSTMbasedEF,Bontemps2017}.

\section{Methods} \label{methods}

The following methods form the core components of an unsupervised anomaly detection approach that uses LSTMs to predict high-volume telemetry data by learning from normal command and telemetry sequences. A novel unsupervised thresholding method is then used to automatically assess hundreds to thousands of diverse streams of telemetry data and determine whether resulting prediction errors represent spacecraft anomalies. Lastly, strategies for mitigating false positive anomalies are outlined and are a key element in developing user trust and improving utility in a production system. 

\subsection{Telemetry Value Prediction with LSTMs}

\textbf{Single-Channel Models.} A single model is created for each telemetry channel and each model is used to predict values for that channel. LSTMs struggle to accurately predict $m$-dimensional outputs when $m$ is large, precluding the input of all telemetry streams into one or a few models. Modeling each channel independently also allows traceability down to the channel level, and low-level anomalies can later be aggregated into various groupings and ultimately subsystems. This enables granular views of spacecraft anomaly patterns that would otherwise be lost. If the system were to be trained to detect anomalies at the subsystem level without this traceability, for example, operations engineers would still need to review a multitude of channels and alarms across the entire subsystem to find the source of the issue. 

Maintaining a single model per channel also facilitates more granular control of the system. Early stopping can be used to limit training to models and channels that show decreases in validation error \cite{caruana2001overfitting}. When issues arise such as high-variance predictions due to overfitting, these issues can be handled on a channel-by-channel basis without affecting the system as a whole.

\begin{figure} 
\includegraphics[height=2.25in]{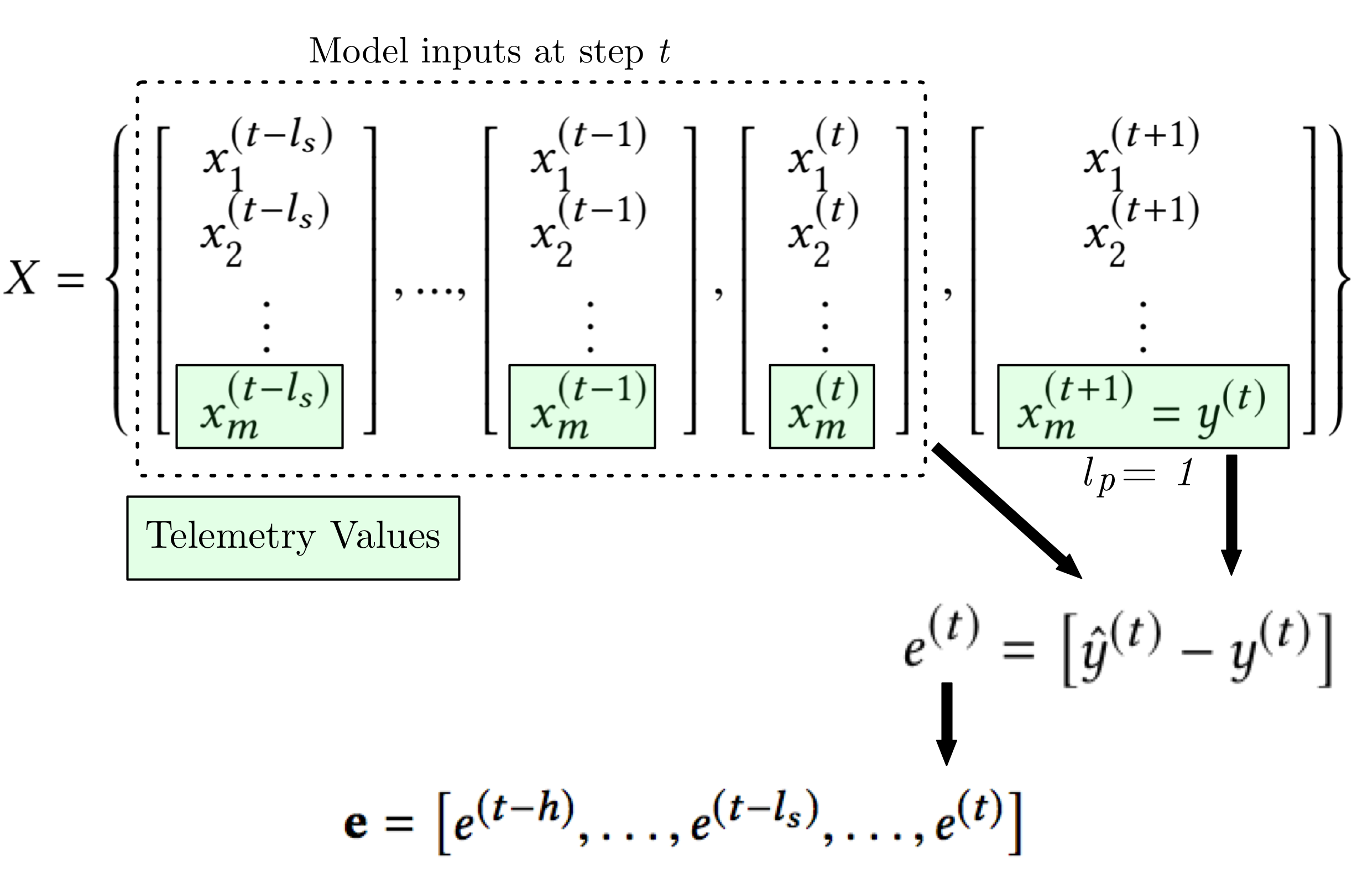}
\caption{A visual representation of the input matrices used for prediction at each time step $t$. Current prediction errors are compared to past errors to determine if they are anomalous.}
\label{matrix}
\end{figure}

\textbf{Predicting Values for a Channel.} Consider a time series \newline $X = \{\textbf{x}^{(1)}, \textbf{x}^{(2)}, \ldots, \textbf{x}^{(n)}\}$ where each step $\textbf{x}^{(t)}\in R^m$ in the time series is an $m$-dimensional vector $\{x_1^{(t)}, x_2^{(t)},\ldots,  x_m^{(t)}\}$, whose elements correspond to input variables \cite{Malhorta2015}. For each point $\textbf{x}^{(t)}$, a sequence length $l_s$ determines the number of points to input into the model for prediction. A prediction length $l_p$ then determines the number of steps ahead to predict, where the number of dimensions $d$ being predicted is $1 \leq d \leq m$. Since our aim is to predict telemetry values for a single channel we consider the situation where $d=1$. We also use $ l_p = 1$ to limit the number of predictions for each step $t$ and decrease processing time. As a result, a single scalar prediction $\hat{y}^{(t)}$ is generated for the actual telemetry value at each step $t$ (see Figure \ref{matrix}). In situations where either $l_p > 1$ or $d > 1$ or both, Gaussian parameters can be used to represent matrices of predicted values at a single step $t$ \cite{Malhorta2015}. 

In our telemetry prediction scenario, the inputs $\textbf{x}^{(t)}$ into the LSTM consist of prior telemetry values for a given channel and encoded command information sent to the spacecraft. Specifically, the combination of the module to which a command was issued and whether a command was sent or received are one-hot encoded and slotted into each step $t$ (see Figure \ref{full-example}). 



\subsection{Dynamic Error Thresholds} \label{error-thresholds}

Automated monitoring of thousands of telemetry channels whose expected values vary according to changing environmental factors and command sequences requires a fast, general, and unsupervised approach for determining if predicted values are anomalous. One common approach is to make Gaussian assumptions about the distributions of past smoothed errors as this allows for fast comparisons between new errors and compact representations of prior ones \cite{ahmad2017unsupervised,Shipmon2017}. However, this approach often becomes problematic when parametric assumptions are violated as we demonstrate in Section 4.3, and we offer an approach that efficiently identifies extreme values without making such assumptions. Distance-based methods are similar in this regard but they often involve high computational cost, such as those that call for comparisons of each point to a set of $k$ neighbors \cite{Gao2012,kriegel2009loop}. Also, these methods are more general and are concerned with anomalies that occur in the normal range of values. Only abnormally high or low smoothed prediction errors are of interest and error thresholding is, in a sense, a simplified version of the initial anomaly detection problem.

\textbf{Errors and Smoothing.} Once a predicted value $\hat{y}^{(t)}$ is generated for each step $t$, the prediction error is calculated as $e^{(t)} = |y^{(t)} - \hat{y}^{(t)}|$, where $y^{(t)} = x_i^{(t+1)}$ with $i$ corresponding to the dimension of the true telemetry value (see Figure \ref{matrix}). Each $e^{(t)}$ is appended to a one-dimensional vector of errors: \[\textbf{e} = [e^{(t-h)}, \ldots, e^{(t-l_s)},\ldots, e^{(t-1)}, e^{(t)}]\] where $h$ is the number of historical error values used to evaluate current errors. The set of errors $\textbf{e}$ are then smoothed to dampen spikes in errors that frequently occur with LSTM-based predictions -- abrupt changes in values are often not perfectly predicted and result in sharp spikes in error values even when this behavior is normal \cite{Shipmon2017}. We use an exponentially-weighted average (EWMA) to generate the smoothed errors $\textbf{e}_s = [e_s^{(t-h)}, \ldots, e_s^{(t-l_s)} ,\ldots, e_s^{(t-1)}, e_s^{(t)}]$ \cite{hunter1986exponentially}. To evaluate whether values are nominal, we set a threshold for their smoothed prediction errors -- values corresponding to smoothed errors above the threshold are classified as anomalies. 

\textbf{Threshold Calculation and Anomaly Scoring.} At this stage, an appropriate  anomaly threshold is sometimes learned with supervised methods that use labeled examples, however it is often the case that sufficient labeled data is not available and this holds true in our scenario \cite{Chandola2009}. We propose an unsupervised method that achieves high performance with low overhead and without the use of labeled data or statistical assumptions about errors. With a threshold $\epsilon$ selected from the set: \[ \boldsymbol{\epsilon} = \mu(\textbf{e}_s) + \textbf{z}\sigma(\textbf{e}_s)\] Where $\epsilon$ is determined by: 
\[ \epsilon = argmax(\boldsymbol{\epsilon}) = \frac{\Delta \mu(\textbf{e}_s) / \mu(\textbf{e}_s)) + ( \Delta \sigma(\textbf{e}_s)/ \sigma(\textbf{e}_s)} {\vert\textbf{e}_a\vert + \vert\textbf{E}_{seq}\vert^2} \] 


Such that: \newline \indent \indent \indent \indent \indent \indent $\Delta \mu(\textbf{e}_s) = \mu(\textbf{e}_s) - \mu(\{e_s \in \textbf{e}_s \vert e_s < \epsilon\})$ \newline \indent \indent \indent \indent \indent \indent  $\Delta \sigma(\textbf{e}_s) = \sigma(\textbf{e}_s) - \sigma(\{e_s \in \textbf{e}_s | e_s < \epsilon\})$ \newline \indent \indent \indent \indent \indent \indent $\textbf{e}_a = \{e_s \in \textbf{e}_s | e_s > \epsilon\}$ \newline \indent \indent \indent \indent \indent \indent $ \textbf{E}_{seq} = $ continuous sequences of $e_a \in \textbf{e}_a$ \newline

Values evaluated for $\epsilon$ are determined using $z \in \textbf{z} $ where $\mathbf{z}$ is an ordered set of positive values representing the number of standard deviations above $\mu(\textbf{e}_s)$. Values for $\mathbf{z}$ depend on context, but we found a range of between two and ten to work well based on our experimental results. Values for $z$ less than two generally resulted in too many false positives. Once $argmax(\boldsymbol{\epsilon})$ is determined, each resulting anomalous sequence of smoothed errors $\textbf{e}_{seq} \in \textbf{E}_{seq}$ is given an anomaly score, $s$, indicating the severity of the anomaly: \[s^{(i)} = \frac{max(\textbf{e}_{seq}^{(i)}) - argmax(\boldsymbol{\epsilon})} {\mu(\textbf{e}_s) + \sigma(\textbf{e}_s)}\]

In simple terms, a threshold is found that, if all values above are removed, would cause the greatest percent decrease in the mean and standard deviation of the smoothed errors $\textbf{e}_s$. The function also penalizes for having larger numbers of anomalous values ($\vert\textbf{e}_a\vert$) and sequences ($\vert\textbf{E}_{seq}\vert$) to prevent overly greedy behavior. Then the highest smoothed error in each sequence of anomalous errors is given a normalized score based on its distance from the chosen threshold.


\subsection{Mitigating False Positives} \label{pruning}
\textbf{Pruning Anomalies.} The precision of prediction-based anomaly detection approaches heavily depends on the amount of historical data ($h$) used to set thresholds and make judgments about current prediction errors. At large scales it becomes expensive to query and process historical data in real-time scenarios and a lack of history can lead to false positives that are only deemed anomalous because of the narrow context in which they are evaluated. Additionally, when extremely high volumes of data are being processed a low false positive rate can still overwhelm human reviewers charged with evaluating potentially anomalous events. 

\begin{figure} 
\includegraphics[height=2.35in]{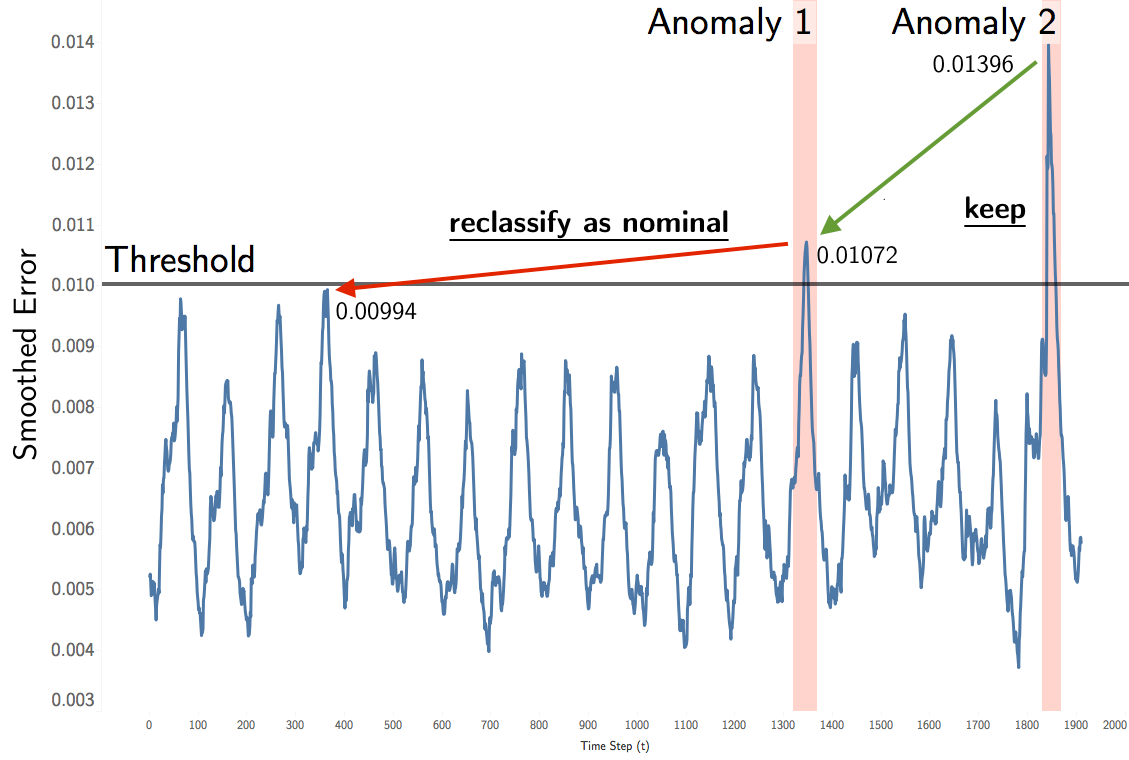}
\caption{An example demonstrating the anomaly pruning process. In this scenario $\textbf{e}_{max} = [0.01396, 0.01072, 0.00994]$ and the minimum percent decrease $p = 0.1$. The decrease from Anomaly 2 to Anomaly 1 $d^{(1)} = 0.23 > p$ and this sequence retains its classification as anomalous. From Anomaly 1 to the next highest smoothed error ($e_s = 0.0099$) $d^{(2)} = .07 < p$ so this sequence is re-classified as nominal.}
\label{pruning-example}
\end{figure}

To mitigate false positives and limit memory and compute cost, we introduce a pruning procedure in which a new set, $\textbf{e}_{max}$, is created containing $max(\textbf{e}_{seq})$ for all $\textbf{e}_{seq}$ sorted in descending order. We also add the maximum smoothed error that isn't anomalous, $max(\{e_s \in \textbf{e}_s \in \textbf{E}_{seq} | e_s \ni \textbf{e}_a\})$, to the end of $\textbf{e}_{max}$. The sequence is then stepped through incrementally and the percent decrease $ d^{(i)} = (e_{max}^{(i-1)} - e_{max}^{(i)}) / e_{max}^{(i-1)}$ at each step $i$ is calculated where $i \in \{1,2,...,(\vert\textbf{E}_{seq}\vert+1)\}$. If at some step $i$ a minimum percentage decrease $p$ is exceeded by $d^{(i)}$, all $e_{max}^{(j)} \in \textbf{e}_{max} |\ j < i$ and their corresponding anomaly sequences remain anomalies. If the minimum decrease $p$ is not met by $d^{(i)}$ \textit{and} for all subsequent errors $d^{(i)}, d^{(i+1)}, \ldots, d^{(i+\vert\textbf{E}_{seq}\vert+1)}$ those smoothed error sequences are reclassified as nominal. This pruning helps ensures anomalous sequences are not the result of regular noise within a stream, and it is enabled through the initial identification of sequences of anomalous values via thresholding. Limiting evaluation to only the maximum errors in a handful of potentially anomalous sequences is much more efficient than the multitude of value-to-value comparisons required without thresholding. 

\textbf{Learning from History.} A second strategy for limiting false positives can be employed once a small amount of anomaly history or labeled data has been gathered. Based on the assumption that anomalies of similar magnitude $s$ generally are not frequently recurring within the same channel, we can set a minimum score, $s_{min}$, such that future anomalies are re-classified as nominal if $s < s_{min}$. A minimum score would only be applied to channels of data for which the system was generating anomalies above a certain rate and $s_{min}$ is individually set for all such channels. Prior anomaly scores for a channel can be used to set an appropriate $s_{min}$ depending on the desired balance between precision and recall. 

Additionally, if the anomaly detection system has a mechanism by which users can provide labels for anomalies, these labels can also be used to set $s_{min}$ for a given stream. For example, if a stream or channel has several confirmed false positive anomalies, $s_{min}$ can be set near the upper bound of these false positive anomaly scores. Both of these approaches have played an important role in improving the precision of early implementations of the system by helping account for normal spacecraft behaviors that are infrequent but occur at regular intervals.


\section{Experiments}

\begin{figure} 

\includegraphics[height=4.58in]{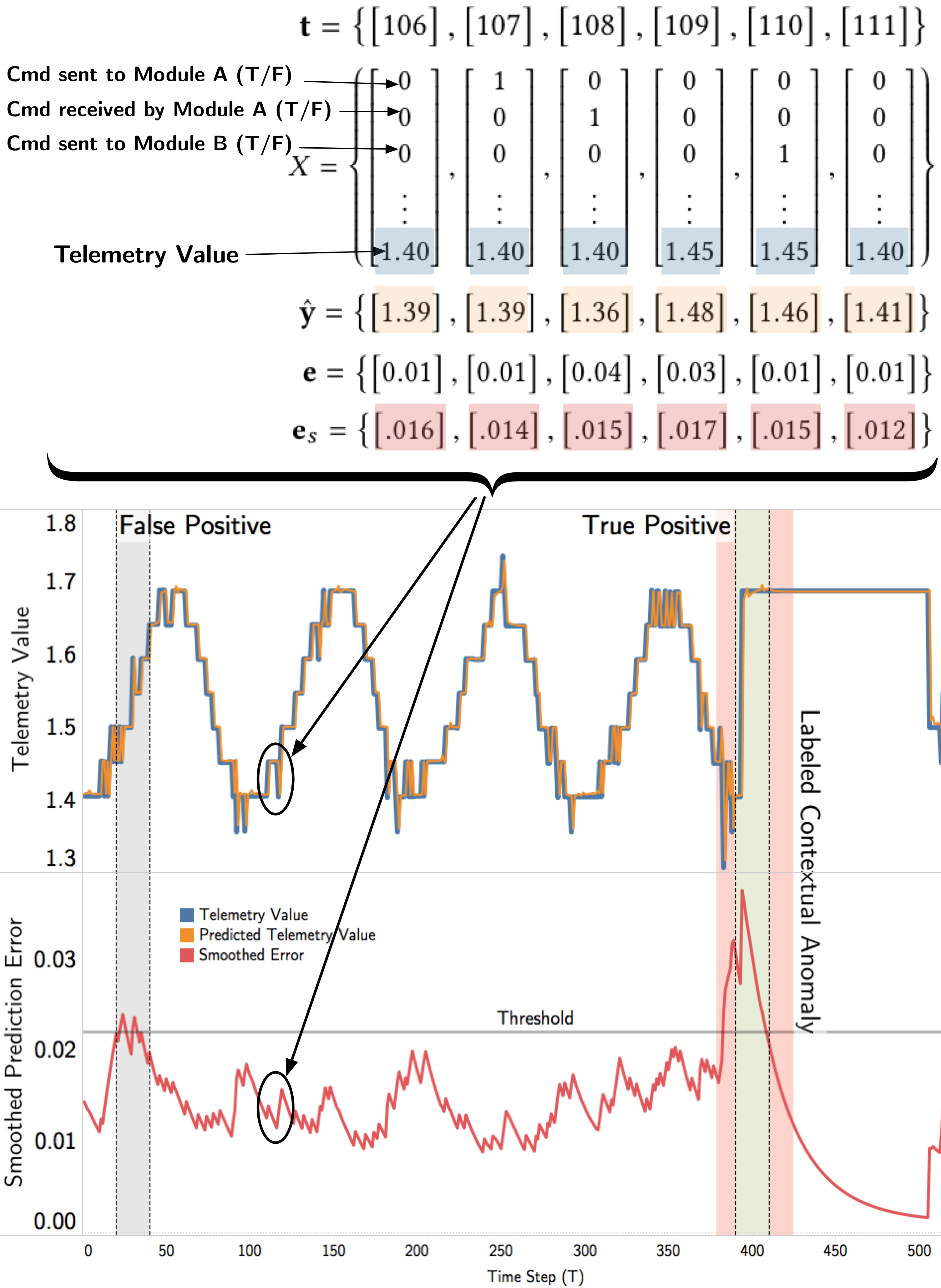}
\caption{The encoding the of command information is demonstrated for a telemetry stream containing a \textit{contextual} anomaly that is unlikely to be identified using limit- or distance-based approaches. Using the encoded command information and prior telemetry values for the channel, predictions are generated for the next time step with resulting errors. The one-step-ahead predictions and actual telemetry values are very close in this example as shown in top time series plot. An error threshold is set using the non-parametric thresholding approach detailed in Section \ref{error-thresholds}, resulting in two predicted anomalous sequences -- one false positive and one true positive lying within the labeled anomalous region. The false positive demonstrates the need for pruning described in Section \ref{pruning}, which would reclassify that sequence as nominal given that it is relatively close to values below the threshold (see Figure \ref{pruning-example}). }
\label{full-example}
\end{figure}

For many spacecraft including SMAP and MSL, current anomaly detection systems are difficult to assess. The precision and recall of alarms aren't captured and telemetry assessments are often performed manually. Fortunately, indications of telemetry anomalies can be found within previously mentioned ISA reports. A subset of all of the incidents and anomalies detailed in ISAs manifest in specific telemetry channels, and by mining the ISA reports for SMAP and MSL we were able to collect a set of telemetry anomalies corresponding to actual spacecraft issues involving various subsystems and channel types.

\begin{table}[htb]
\centering
\caption{Experimental Data Information}
\label{experiment-data}
\begin{tabular}{@{}|l|c|c|c|@{}}
\toprule
                               & \textbf{SMAP}      & \textbf{MSL}       & \textbf{Total} \\ \midrule
Total anomaly sequences           & 69        & 36        & \textbf{105}       \\
\textit{Point} anomalies (\% tot.)     & 43 (62\%) & 19 (53\%) & \textbf{62 (59\%)}      \\
\textit{Contextual} anomalies (\% tot.) & 26 (38\%) & 17 (47\%) & \textbf{43 (41\%)}    \\
Unique telemetry channels       & 55        & 27        & \textbf{82}             \\
Unique ISAs          & 28     & 19     & \textbf{47}          \\
Telemetry values evaluated          & 429,735     & 66,709     & \textbf{496,444}          \\ \bottomrule
\end{tabular}
\end{table}


All telemetry channels discussed in an individual ISA were reviewed to ensure that the anomaly was evident in the associated telemetry data, and specific anomalous time ranges were manually labeled for each channel. If multiple anomalous sequences and channels closely resembled each other, only one was kept for the experiment in order to create a diverse and balanced set.


We classify anomalies into two categories, \textit{point} and \textit{contextual}, to distinguish between anomalies that would likely be identified by properly-set alarms or distance-based methods that ignore temporal information (\textit{point} anomalies) and those that require more complex methodologies such as LSTMs or Hierarchical Temporal Memory (HTM) approaches to detect (\textit{contextual} anomalies)\cite{ahmad2017unsupervised}. This characterization is adapted from the three categories previously mentioned -- \textit{point}, \textit{contextual}, and \textit{collective} \cite{Chandola2009}. Since \textit{contextual} and \textit{collective} anomalies both require temporal context and are harder to detect, they have both been combined into the \textit{contextual} category presented in the next section.

In addition to evaluating the performance of the methodologies in Section \ref{methods}, we also compare the post-prediction performance of our error thresholding method to a parametric unsupervised approach used in the top-performing algorithm for the recent Numenta Anomaly Benchmark \cite{ahmad2017unsupervised, lavin2015evaluating}. 


No comparisons are made between the LSTM-based approach and other predictive models as leaps forward in the underlying prediction performance will more likely come from providing increasingly refined command-based features to the model. Given the rise in prediction-based anomaly detection methods and related research \cite{Malhotra2016LSTMbasedEF,Malhorta2015}, we place increased emphasis on post-prediction error evaluation methods that have received comparatively less focus yet demonstrate significant impact on our results.


\subsection{Setup}
For each unique stream of data containing one or more anomalous sequences with the primary anomaly occurring at time $t_a$, we evaluate all telemetry values in a surrounding timeframe from $t_{s} = t_a - 3d$ to $t_{f} = t_a + 2d$ where $d$ is days. A model is trained for each unique stream using values and command data from $t_{s_{train}} = t_s - 2d$ to $t_{f_{train}} = t_s$. Additional days were included if sufficient data was not available in these timeframes. This 5-day span around anomalies was selected to balance two objectives: deeper insight into precision and reasonable computational cost. Predicted anomalous regions are also slightly expanded to facilitate the combining of anomalous regions in close proximity -- regions that overlap or touch after expansion are combined into a single region to account for situations where multiple anomalous regions represent a single event. 

Each labeled anomalous sequence $x_{a} \in \textbf{x}_{a}$ of telemetry values is evaluated against the final set of predicted anomalous sequences identified by the system according to the following rules:
\begin{enumerate}
	\item  A \textbf{true positive} is recorded if: \[|e_{a}^{(t)} \in e_{seq} \in \textbf{e}_{seq}: x_{i}^{(t)} \in x_a| > 0\] for any $x_{a} \in \textbf{x}_{a}$. In other words, a true positive results if any portion of a predicted sequence of anomalies falls within any true labeled sequence. Only one true positive is recorded even if portions of multiple predicted sequences fall within a labeled sequence.
    \item If no predicted sequences overlap with a positively labeled sequence, a \textbf{false negative} is recorded for the labeled sequence.
    \item For all predicted sequences that do not overlap a labeled anomalous region, a \textbf{false positive} is recorded. 
\end{enumerate}

For simplicity, we do not make scoring adjustments based on how early an anomaly was detected or the distance between false positives and labeled regions \cite{lavin2015evaluating}.

\textbf{Batch processing.} Telemetry values are aggregated into one minute windows and evaluated in batches of 70 minutes mimicking the downlink schedule for SMAP and our current system implementation. Each 70 minute batch of values is evaluated using $h=2100$, where $h$ is the number of prior values used to calculate an error threshold and evaluate the current batch. The system is also well-suited to process values in a real-time, streaming fashion when applicable.


\subsection{Model Parameters and Evaluation}

The same architecture and parameters are used for all models in the experiment:

\begin{table}[htb]
\centering
\label{my-label}
\begin{tabular}{lc}
\multicolumn{2}{c}{\textbf{Model Parameters}}    \\ \hline \\[-1em] 
hidden layers              & 2                   \\
units in hidden layers & 80                  \\
sequence length ($l_s$)    & 250                 \\
training iterations        & 35                  \\
dropout                    & 0.3                 \\
batch size                 & 64                  \\
optimizer                  & Adam                \\
input dimensions           & 25 (SMAP), 55 (MSL) \\ \hline
\end{tabular}
\end{table}

Each model is shallow with only two hidden layers and 80 units in each layer. We found this architecture provided enough capacity to predict individual channels well, and adding additional capacity provided little to no prediction benefits while increasing model sizes and training times. All channels do not necessarily require this amount of capacity and future improvements could include automated selection of appropriate model capacity based on channel complexity. Similarly, a sequence length $l_s = 250$ provided a balance between performance and training times. The difference in input dimensions for SMAP and MSL results from the missions each having different sets of command modules. Early stopping was used to prevent overfitting during model training, and not all models were trained for the full 35 iterations. 

Once predictions were generated, anomaly thresholds for smoothed errors were calculated using the method detailed in Section \ref{error-thresholds} with $\textbf{z}=\{2.5, 3.0, 3.5, ... , 10.0\}$ and the minimum percent difference between subsequent anomalies $p=0.13$. The $p$ parameter is an important lever for controlling precision and recall, and an appropriate value can be inferred when labels are available. In our setting, reasonable results were achieved with $0.05 < p < 0.20$ (see Figure \ref{precision-recall}).

\textbf{Comparison with Parametric Thresholding.} Using the raw LSTM prediction errors, we also generate anomalies with the parametric error evaluation approach used in coordination with the most accurate model from the Numenta Anomaly Benchmark \cite{lavin2015evaluating}. This approach processes raw errors incrementally -- at each step $t$ a window $W$ of historical errors is modeled as a normal distribution, and the mean $\mu_W$ and variance $\sigma_{W}^2$ are updated at each step $t$. We set $W$'s length $l_w = h = 2100$ and use the same set of prediction errors for both approaches. A short-term average $\mu_s$ of length $l_{short}$ of prediction errors is then calculated and has a similar smoothing effect as the EWMA smoothing in Section \ref{error-thresholds}. The likelihood of an anomaly $L$ is then defined using the tail probability $Q$: \[L = 1 - Q\left(\frac{\mu_s - \mu_W}{\sigma_{W}^2}\right)\]

If $L \geq 1 - \epsilon_{norm}$ values are classified as anomalous. In the next section, results generated using $l_{short} = 10$ and $\epsilon_{norm} = \{0.01, 0.0001\}$ are compared to the approach in Section \ref{error-thresholds}. The effects of pruning (detailed in Section \ref{pruning}) on this approach are also tested.

\subsection{Results and Discussion} \label{results}

\begin{table}[]
\centering
\caption{Results for each spacecraft using LSTM predictions and various approaches to error thresholding.}
\label{results}
\begin{tabular}{@{}rccc@{}}
\toprule
\multicolumn{1}{l}{Thresholding Approach} & \multicolumn{1}{r}{Precision} & \multicolumn{1}{l}{Recall} & \multicolumn{1}{l}{$F_{0.5}$ score} \\ \midrule
\multicolumn{4}{l}{\textbf{Non-Parametric  w/ Pruning ($p = 0.13$)}} \\[-1em]                                   & \multicolumn{1}{r}{}                & \multicolumn{1}{r}{}                 \\
MSL                                                & 92.6\%                                 & 69.4\%                              & \textbf{0.69}                                 \\
SMAP                                               & 85.5\%                                 & 85.5\%                              & \textbf{0.71}                                 \\
Total                                              & 87.5\%                                 & 80.0\%                              & \textbf{0.71}                        \\  \\[-1em]
\multicolumn{4}{l}{\textbf{Non-Parametric w/out Pruning ($p = 0$)}} \\[-1em]                                &                                     &                                      \\ 
MSL                                                & 75.8\%                                 & 69.4\%                              & 0.61                                 \\
SMAP                                               & 43.0\%                                 & 92.8\%                              & 0.44                                 \\
Total                                              & 48.9\%                                 & 84.8\%                    & 0.47                                 \\ \\[-1em]
\multicolumn{4}{l}{\textbf{Gaussian Tail ($\epsilon_{norm} = 0.0001$)}} \\[-1em]                                     &                                     &                                      \\
MSL                                                & 84.2\%                                 & 44.4\%                              & 0.54                                 \\
SMAP                                               & 88.5\%                                 & 78.3\%                              & 0.71                                 \\
Total                                              & 87.5\%                                 & 66.7\%                              & 0.66                                 \\ \\[-1em]
\multicolumn{4}{l}{\textbf{Gaussian Tail ($\epsilon_{norm} = 0.01$)}} \\[-1em]                                    &                                     &                                      \\
MSL                                                & 61.3\%                                 & 52.8\%                              & 0.48                                 \\
SMAP                                               & 82.4\%                                 & 81.2\%                              & 0.68                                 \\
Total                                              & 75.8\%                                 & 71.4\%                              & 0.62                                 \\ \\[-1em]
\multicolumn{4}{l}{\textbf{Gaussian Tail w/ Pruning ($\epsilon_{norm} = 0.01, p = 0.13$)}} \\[-1em]                          &                                     &                                      \\ 
MSL                                                & 88.2\%                                 & 41.7\%                              & 0.54                                 \\
SMAP                                               & 92.7\%                                 & 73.9\%                              & 0.71                                 \\
Total                                              & 91.7\%                        & 62.9\%                              & 0.66                                 \\ \\[-1em] \bottomrule
\end{tabular}
\end{table}

As shown in Table 2, the best results in terms of $F_{0.5}$ score are achieved using the LSTM-based predictions combined with the non-parametric thresholding approach with pruning. In terms of prediction, The LSTM models achieved an average normalized absolute error of 5.9\% predicting telemetry values one time step ahead. 

\begin{table}[htb]
\centering
\caption{Telemetry Prediction Errors}
\label{predictions}
\begin{tabular}{@{}rlrl@{}}
\toprule
\multicolumn{4}{l}{\textbf{Average LSTM Prediction Error}}              \\ \midrule
\multicolumn{2}{c}{\phantom{move!} MSL}            & \multicolumn{2}{c}{6.8\%}          \\
\multicolumn{2}{c}{\phantom{move !}SMAP}           & \multicolumn{2}{c}{5.5\%}          \\
\multicolumn{2}{c}{\phantom{move!}\textbf{Total}} & \multicolumn{2}{c}{\textbf{5.9\%}} \\ \bottomrule
\end{tabular}
\end{table}

Parameters were tuned to balance precision and recall for experimentation, however in the current implementation precision is weighted more heavily when tuning parameters because the precision results shown are overly optimistic compared to the actual implementation of the system. There is an implicit assumption in the experiment that anomalies occur once every five days, where five days is the total number of days processed for each stream containing an anomaly. The experiment also does not include processing for all streams not exhibiting anomalous behavior for a given time window, which would further increase the number of false positives. This decreased precision in the implemented system is offset by setting minimum anomaly scores via the methods outlined at the end of Section \ref{pruning}. 

\begin{figure} 
\includegraphics[height=3.35in]{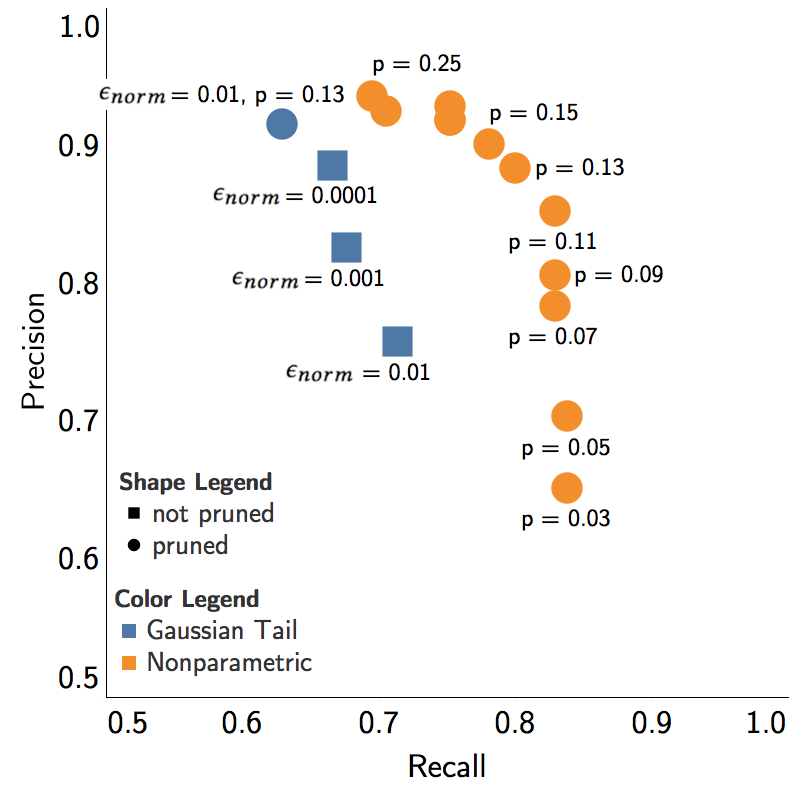}
\caption{Plot showing comparison of overall precision and recall results for parametric approach and approach presented in this paper (labeled 'Nonparametric') with various parameter settings.}
\label{precision-recall}
\end{figure}

\textbf{Thresholding Comparisons.}  Results for the non-parametric approach without pruning are presented to demonstrate pruning's importance in mitigating false positives. The pruning process is roughly analogous to the pruning of decision trees in the sense that it helps pare down a greedy approach designed to overfit in order to improve performance. In this instance, pruning only decreases overall recall by 4.8 percentage points (84.8\% to 80.0\%) while increasing overall precision by 38.6 percentage points (48.9\% to 87.5\%). The 84.8\% recall achieved without pruning is an approximation of the upper bound for recall given the predictions generated by the LSTM models. If predictions are poor and resulting smoothed errors do not contain a signal then thresholding methods will be ineffective.

The Gaussian tail approach results in lower levels of precision and recall using various parameter settings. Pruning greatly improves precision but at a high recall cost, resulting in an $F_{0.5}$ score that is still well below the score achieved by the non-parametric approach with pruning. One factor that contributes to lower performance for this method is the violation of Gaussian assumptions in the smoothed errors. Using D'Agostino and Pearson's normality test \cite{d1973tests}, we reject the null hypothesis of normality for \textit{all} sets of smoothed errors using a threshold of $\alpha = 0.005$. The error information lost when using Gaussian parameters results in suboptimal thresholds that negatively affect precision and recall and cannot be corrected by pruning (see Figure \ref{precision-recall} and Table 2).

\textbf{Performance for Different Anomaly Types.} The high proportion of \textit{contextual} anomalies (41\%) provides further justification for the use of LSTMs and prediction-based methods over methods that ignore temporal information. Only a small subset of the \textit{contextual} anomalies -- those where anomalous telemetry values happen to fall in low-density regions -- could theoretically be detected using limit-based or density-based approaches. Optimistically, this establishes a maximum possible recall near the best result presented here and obviates extensive comparisons with these approaches. Not surprisingly, recall was lower for \textit{contextual} anomalies but the LSTM-based approach was able to identify a majority of these. 

\begin{table}[htb]
\centering
\caption{Recall for different anomaly types using LSTM predictions with non-parametric thresholding and pruning.}
\label{my-label}
\begin{tabular}{@{}rcc@{}}
\toprule
\multicolumn{1}{l}{\textbf{}} & \multicolumn{1}{l}{Recall - \textit{point}} & \multicolumn{1}{l}{Recall - \textit{contextual} } \\ \midrule
MSL                           & 78.9\%                                                     & 58.8\%                                                \\
SMAP                          & 95.3\%                                                     & 76.0\%                                                \\
Total                         & 90.3\%                                                     & 69.0\%                                               
\end{tabular}
\end{table}

\textbf{Performance for Different Spacecraft.} SMAP and MSL are very different missions representing varying degrees of difficulty when it comes to anomaly detection. Compared to MSL, operations for the SMAP spacecraft are routine and resulting telemetry can be more easily predicted with less training and less data. MSL performs a much wider variety of behaviors with varying regularity, some of which resulted during rover activities that were not present in the limited training data. This explains the lower precision and recall performance for MSL ISAs and is also apparent in the difference between the average LSTM prediction errors - average error in predicting telemetry for SMAP was 5.5\% versus 6.8\% for MSL (see Table \ref{predictions}). 

\section{Deployment}



The methods presented in this paper have been implemented into a system that is currently being piloted by SMAP operations engineers.  Over 700 channels are being monitored in near real-time as data is downlinked from the spacecraft and models are trained offline every three days with early stopping. We have successfully identified several confirmed anomalies since the initial deployment in October 2017. However, one major obstacle to becoming a central component of the telemetry review process is the current rate of false positives. High demands are placed on operations engineers and they are hesitant to alter effective procedures. Adopting new technologies and systems means increased risk of wasting valuable time and attention. Investigation of even a couple false positives can deter users and therefore achieving high  precision with over a million telemetry values being processed per day is essential for adoption. 

\textbf{Future Work.} The pilot deployment and experimental results are key milestones in establishing that a large-scale, automated telemetry monitoring system is feasible. Future work will be focused around improving telemetry predictions primarily through improved feature engineering. 

Spacecraft command information is only one-hot encoded at the module level in the current implementation, and no information about the nature of the command itself is passed to the models. Much more granular information around command activity and other sources of information like event records may be necessary to accurately predict telemetry data for missions without routine operations. For these missions, training data from periods with similar activities to those planned must be automatically identified and selected rather than simply training on recent activity. Accurate predictions are critical to this approach and will allow the system to be extended to missions like MSL while also addressing the need for improved precision. The two aforementioned improvements represent key areas of future work that will be generally beneficial for monitoring dynamic and complex spacecraft. We also plan to continue to refine our approaches to mitigating false positives described in Section \ref{pruning} and improve interfaces facilitating the review, investigation, and expert labeling of anomalies found by the system.  

Lastly, another key aspect of our problem that has not been addressed are the interactions and dependencies inherent in the telemetry channels. This has been partially addressed through a visual interface, but a more mathematical and automated view into the correlations between channel anomalies would provide important insight into complex system behaviors and anomalies. 

\section{Conclusion}

This paper presents and defines an important and growing challenge within spacecraft operations that stands to greatly benefit from modern anomaly detection approaches. We demonstrate the viability of LSTMs for predicting spacecraft telemetry while addressing key challenges involving interpretability, scale, precision, and complexity that are inherent in many anomaly detection scenarios. We also propose a novel dynamic thresholding approach that does not rely on scarce labels or false parametric assumptions. Key areas for improvement and further evaluation have also been identified as we look to expand capabilities and implement systems for a variety of spacecraft. Finally, we make public a large real-world, expert-labeled set of anomalous spacecraft telemetry data and offer open-source implementations of the methodologies presented in this paper.

\begin{acks}
This effort was supported by the Office of the Chief Information Officer (OCIO) at JPL, managed by the California Institute of Technology on behalf of NASA. The authors would specifically like to thank Sonny Koliwad, Chris Ballard, Prashanth Pandian, Chris Swan, and Charles Kirby for their feedback and support.

\end{acks}

\bibliographystyle{ACM-Reference-Format}
\balance
\bibliography{refs} 

\end{document}